\documentclass[letterpaper,10pt,conference]{ieeeconf}

\IEEEoverridecommandlockouts
\usepackage{amsmath}
\usepackage{amssymb}
\usepackage{bm}
\usepackage{booktabs}
\usepackage{array}
\usepackage{graphicx}
\usepackage{subcaption}
\usepackage{placeins}
\usepackage{stfloats}
\usepackage{xcolor}
\usepackage{cite}
\usepackage{balance}
\usepackage[hidelinks]{hyperref}

\graphicspath{{figures/}}
\newcommand{\cmark}{\textcolor{green!55!black}{\checkmark}}
\newcommand{\xmark}{\textcolor{red!70!black}{$\times$}}

\title{HANDOFF: Humanoid Agentic Task-Space Whole-Body Control via Distilled Complementary Teachers}

\author{Lizhi Yang$^{1}$, Junheng Li$^{1}$, Nehar Poddar$^{2}$,
Yiling Hou$^{1}$, Gio Huh$^{1}$, \\ Robert Griffin$^{2}$,
Georgia Gkioxari$^{1}$, and Aaron D. Ames$^{1}$%
\thanks{$^{1}$California Institute of Technology,
Pasadena, CA, USA. {\tt\small \{lzyang, junhengl, yhou, ghuh,
georgia, ames\}@caltech.edu}}%
\thanks{$^{2}$The Institute for Human \& Machine Cognition,
Pensacola, FL, USA. {\tt\small \{npoddar, rgriffin\}@ihmc.org}}%
}

\begin{document}
\maketitle
\thispagestyle{empty}
\pagestyle{empty}

\begin{abstract}
Humanoid loco-manipulation benefits from one controller to coordinate locomotion, arm movements, and fall recovery. 
These behaviors are difficult to learn together from scratch because they require different skills and objectives.
We present HANDOFF, a training architecture that distills motion-tracking, locomotion, and recovery teachers into one 29-DoF policy, using context-conditioned KL losses over separate action slices and a soft mixture of experts.
Commanded velocity continuously blends body supervision, while a binary recovery flag assigns the full action to the recovery teacher. 
The resulting controller takes a compact 10-D task-space command rather than a dense kinematic stream and does not switch policies at runtime.
On the Unitree G1, HANDOFF matches state-of-the-art velocity tracking and obtains the largest robust manipulation workspace in a matched adapted-interface benchmark. 
The same controller executes natural-language-driven, multi-stage loco-manipulation tasks in simulation and hardware experiments without further data collection or fine-tuning.
\end{abstract}

\begin{table*}[t]
\centering
\caption{\textbf{Planner-facing interfaces.} Recent whole-body humanoid controllers are compared by their required command references and whether they retain a single deployed policy.}
\label{tab:controller-interfaces}
\footnotesize
\setlength{\tabcolsep}{6pt}
\renewcommand{\arraystretch}{1.15}
\resizebox{\textwidth}{!}{%
\begin{tabular}{l l l l c c c}
\toprule
\textbf{Method} & \textbf{Locomotion ref.} & \textbf{Height ref.} & \textbf{Arm ref.} & \textbf{No dense} & \textbf{No ext.} & \textbf{Single} \\
                &                          &                     &                   & \textbf{streaming} & \textbf{kin. ref.} & \textbf{policy} \\
\midrule
Motion-tracking WBCs~\cite{sonic2025, beyondmimic2025, ze2025twist2,humanplus2024, exbody2_2024, gmt2025} & kin. ref. (29D) & from kin. & full upper kin. (14D) & \xmark & \xmark & \cmark \\
ResMimic~\cite{resmimic2025}        & kin. ref. (29D)  & from kin.   & full upper kin. (14D)    & \xmark & \xmark & \xmark \\
VisualMimic~\cite{visualmimic2025}  & keypoint stream (18D) & from kin.   & keypoint stream (18D)   & \xmark & \cmark & \cmark \\
HOVER~\cite{hover2025}              & multi-mode kin. ($\sim$80D + mask)  & from kin.   & from kin.    & \xmark & \xmark & \cmark \\
HOMIE / FALCON~\cite{homie2025, falcon2026} & velocity (2--6D) & scalar $h$ (1D) & full upper kin. (14D) & \cmark & \xmark & \xmark \\
AMO~\cite{amo2025}                  & velocity (3D)        & torso pose (4D)  & full upper kin. (14D)       & \cmark & \xmark & \xmark \\
HUMI~\cite{humi2026}                & pelvis traj. (6D)  & from pelvis & EE traj. (12D)        & \xmark & \cmark & \cmark \\
HERO~\cite{hero2026}                & velocity (6D)      & scalar $h$ (1D)  & upper kin. (17D) + IK residual (6D)       & \cmark & \xmark & \xmark \\
\midrule
\textbf{HANDOFF (Ours)}              & \textbf{velocity (3D)} & \textbf{scalar $\boldsymbol{h}$ (1D)} & \textbf{wrist target (6D)} & \cmark & \cmark & \cmark \\
\bottomrule
\end{tabular}}
\end{table*}

\section{Introduction}
A humanoid robot fetching a cup of coffee for someone looks effortless in a video. 
The planning and control problem behind that loco-manipulation action is anything but: it has to coordinate object localization, sensing, motion planning, and whole-body control~\cite{gu2026humanoid}.
For the planner, the practical question is what to command.
A person can describe sparse subgoals (``find the coffee, walk near it, reach for it, grab it'') and leave gait and balance to lower-level control. 
Most whole-body controllers (WBCs) instead operate in a motion-tracking regime~\cite{sonic2025, beyondmimic2025, ze2025twist2} and require a dense full-body kinematic stream at the control rate.
Producing this stream for a new skill requires teleoperation or motion capture, embodiment-specific retargeting, and dynamic-feasibility filtering. As a result, the planner largely replays motions from a fixed demonstration library.
We seek a low-dimensional command that a human, geometric planner, or VLM can provide while leaving whole-body coordination to the controller.

\begin{figure}[t]
  \centering
  \includegraphics[width=\columnwidth]{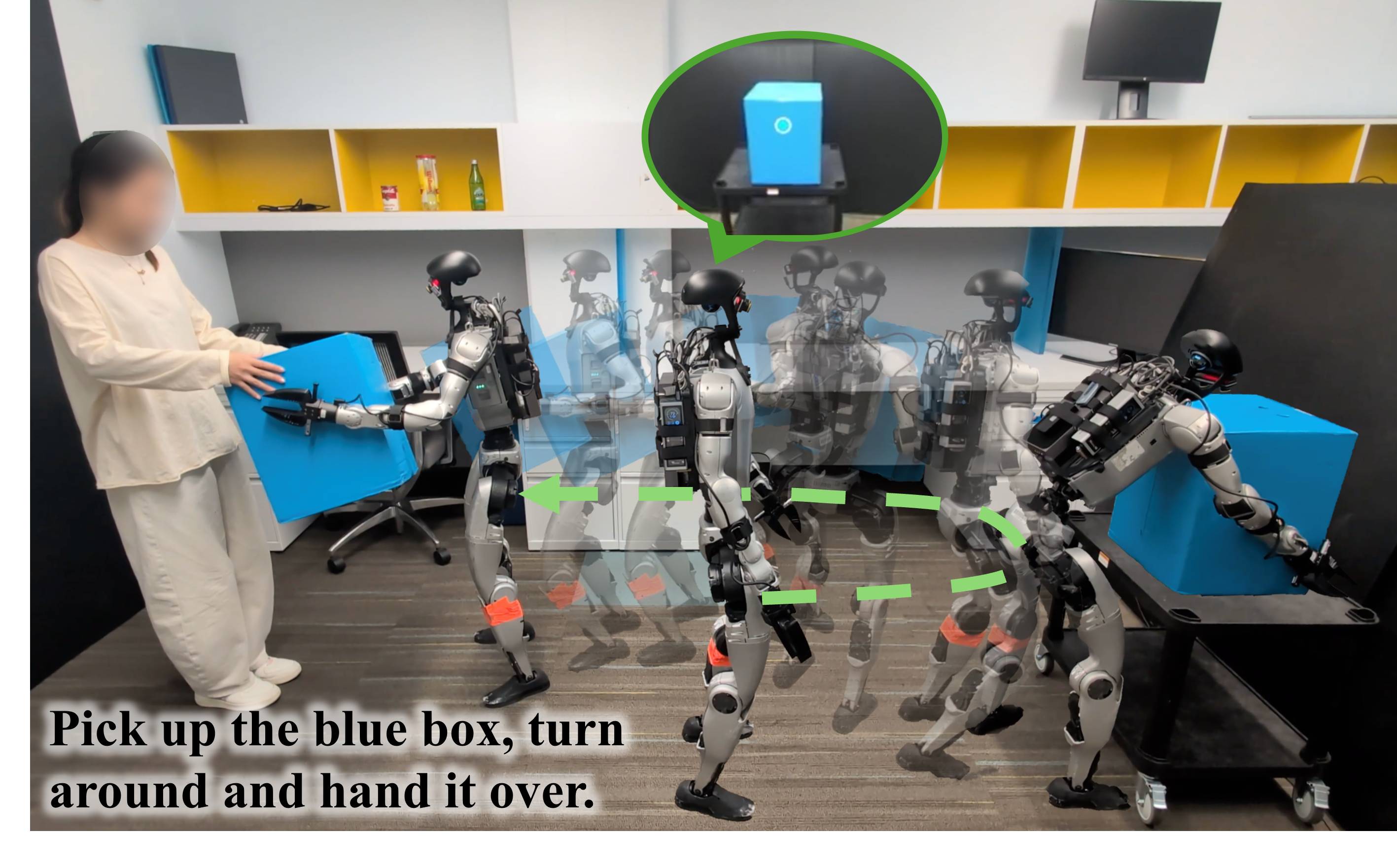}
  \caption{\textbf{HANDOFF.} Motion-tracking, locomotion, and recovery specialists are distilled into one whole-body controller. Context determines which teacher supervises each action slice, and a VLM-based planner drives the resulting 10-D interface without task-specific fine-tuning.}
  \label{fig:teaser}
  \vspace{-10pt}
\end{figure}

In this paper, we propose HANDOFF, a training architecture that composes complementary specialists into one task-space whole-body controller with the command
\[
c_t = \bigl[\,v_x,\ v_y,\ \omega_z,\ z,\ p_L^P,\ p_R^P\,\bigr],
\]
where $v_x, v_y, \omega_z$ are planar base velocity commands, $z$ is the commanded root height, and $p_L^P, p_R^P$ are bilateral pelvis-frame wrist targets.
These variables are standard planner outputs. 
Locomotion stacks produce $(v_x,v_y,\omega_z)$, grasp planners produce end-effector targets, and a squat or reach heuristic can set $z$.
Language-grounded planners~\cite{saycan2022, palme2023} also decompose long-horizon goals into task-space subgoals. VLAs~\cite{rt22023, openvla2024} produce end-effector actions that map to the wrist targets without retargeting or controller fine-tuning.
The command also specifies coupled whole-body behavior.
A low $z$ with forward wrist targets produces a squat and reach; an asymmetric wrist target during locomotion produces a one-handed moving reach.

No single training regime learns velocity tracking, coordinated whole-body reaching, and fall recovery equally well.
We therefore train one specialist for each regime and distill them into a single student.
A public retargeted motion dataset supplies the whole-body motion-tracking teacher~\cite{ze2025twist2,bones_seed_2026}, task rewards train the locomotion teacher~\cite{yang2025cbf}, and adversarial motion priors train the fall-recovery teacher~\cite{peng2021amp, amp_mjlab}.
Each teacher is a separate module that can be improved or replaced independently.
We train the student with PPO~\cite{schulman2017proximal}, multi-teacher KL~\cite{hinton2015distilling}, and a soft mixture of experts~\cite{shazeer2017moe}.
Its objective blends WBC and locomotion supervision over the legs and waist according to commanded velocity, keeps the arms under the motion-tracking teacher, and uses full-action recovery supervision only during recovery. 
This produces one policy with all three behaviors and no runtime switching.

Our contributions are as follows: 
1) a context-conditioned, action-sliced KL objective for distilling teachers with different action coverage into one MoE student;
2) a 29-DoF controller combining locomotion, whole-body coordination, and recovery under one 10-D command without policy switching; and 
3) ablations and matched evaluations of controller interfaces, recovery, randomized tasks, and hardware deployment.
We will release the full framework to facilitate future research in the field.

\section{Related Work}

\noindent\textbf{Whole-body motion tracking with dense-reference interfaces:}
Motion imitation first produced diverse behavior for physics-based characters~\cite{peng2021amp,peng2018deepmimic} and now scales to humanoid robots through motion-tracking WBCs~\cite{sonic2025, beyondmimic2025, ze2025twist2,humanplus2024, exbody2_2024, gmt2025} and masked multi-mode policies~\cite{hover2025}.
Recent work studies residual control~\cite{resmimic2025}, visual hierarchies~\cite{visualmimic2025}, motion data~\cite{opt2skill2024, omniretarget2025}, and real-time retargeting from human demonstrations~\cite{penco2018robust, araujo2025retargeting}.
These controllers reproduce expressive motion, but the planner must generate the full kinematic stream, as in Table~\ref{tab:controller-interfaces}. 
HANDOFF does not reproduce motion outside its teacher coverage, such as dancing, but its commands can be generated by a simple planner.

\noindent\textbf{Task-command controllers and architectural decompositions:}
Controllers with compact mid-level commands differ in how they divide control over the body.
\cite{homie2025} uses cockpit commands for the upper body and a separate RL policy for the lower body. 
\cite{falcon2026} co-trains lower-body locomotion with a force-adaptive upper-body controller, while \cite{amo2025} combines RL with dynamically optimized whole-body references.
\cite{omnih2o2024} and \cite{hover2025} use head-and-hand targets, but still require dense target trajectories at the controller rate.
Other methods consume upper-body joint references~\cite{hero2026}, select from skill libraries~\cite{r2s22025}, or train a task policy above a learned controller~\cite{demohlm2025, humi2026, hitter2025, boxlocomanip2023}. 
In multi-task loco-manipulation systems~\cite{demohlm2025, humi2026}, a new behavior requires new whole-body demonstrations and a new task policy.
Table~\ref{tab:controller-interfaces} summarizes the references exposed to an upstream planner. 
Here, \emph{No ext.\ kin.\ ref.} means that the planner can issue commands without using IK or retargeting to generate joint angles. 
HANDOFF directly admits base velocity, root height, and bilateral wrist-position targets through a single policy.

\noindent\textbf{VLA- and VLM-driven systems:}
Planner-controller systems couple language models to skill libraries~\cite{saycan2022}, while vision-language-action models train transformers to produce low-level actions~\cite{palme2023, rt22023, openvla2024}.
Humanoid systems pre-train on egocentric video and post-train an action expert above a low-level controller~\cite{wholebodyvla2026, psi02026}. 
\cite{being02025} connects a VLM to a modular RL/ACT skill library, and \cite{leverb2026} uses a latent ``verb'' vocabulary instead of an explicit interface.
In HANDOFF, this layer is replaceable and targets the same 10-D base-and-hand command.

\noindent\textbf{Specialist-to-generalist distillation and expert routing:}
Several humanoid systems combine control modes, locomotion behaviors, motion clusters, embodiment-specific specialists, parkour skills, or motion-tracking teachers in one student~\cite{hover2025, ahc2025, bumblebee2025, eagle2026, php2026}.
\cite{gmt2025} gates a motion MoE over clusters from one motion-tracking manifold and retains a dense kinematic reference.
\cite{telegate2026} instead keeps the experts separate and learns an inference-time gate.
\cite{tessler2024maskedmimic} uses DAgger to distill one dense-reference motion-tracking teacher into a partial-observation student for simulated characters, addressing a difference in input modalities.
HANDOFF addresses a different mismatch. Under the fixed command $[v_x,v_y,\omega_z,z,p_L^P,p_R^P]$, the human-motion teacher provides whole-body coordination but limited task coverage, while the locomotion and recovery teachers provide narrower but more reliable behaviors.

\begin{figure*}[t]
\centering
\includegraphics[width=\textwidth]{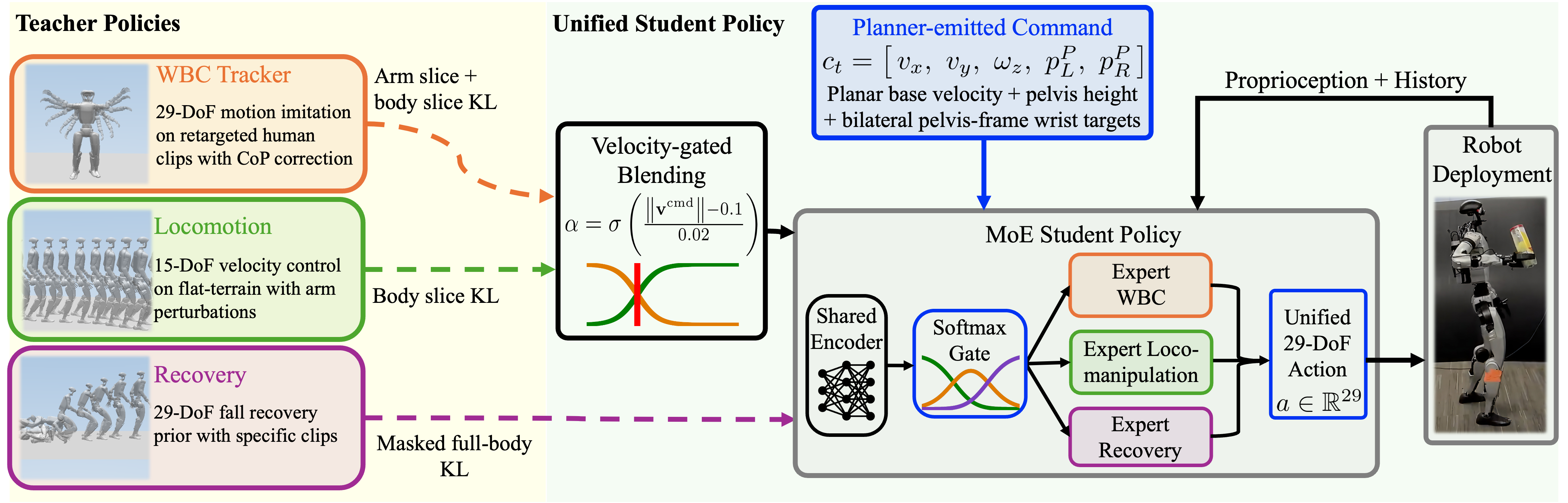}
\caption{\textbf{Training architecture.} Separate whole-body motion-tracking, locomotion, and fall-recovery teachers supervise a context-routed MoE student that maps the 10-D command to 29-DoF actions.}
\label{fig:system}
\end{figure*}

\section{Whole-Body Control}
\label{sec:wbc}
\vspace{-5pt}
Figure~\ref{fig:system} summarizes the training pipeline.
Whole-body coordination, command-conditioned locomotion, and ground-to-upright recovery rely on different references, rewards, and state distributions.
We train one PPO teacher for each regime, then distill them into a single student to avoid runtime policy switching.
Action-sliced, context-conditioned supervision and a soft MoE preserve the specialists' roles within one deployed controller.
\subsection{Whole-Body Control Teacher}
\label{sec:wbc-teacher}

\begin{figure}[!t]
\centering
\includegraphics[width=\columnwidth]{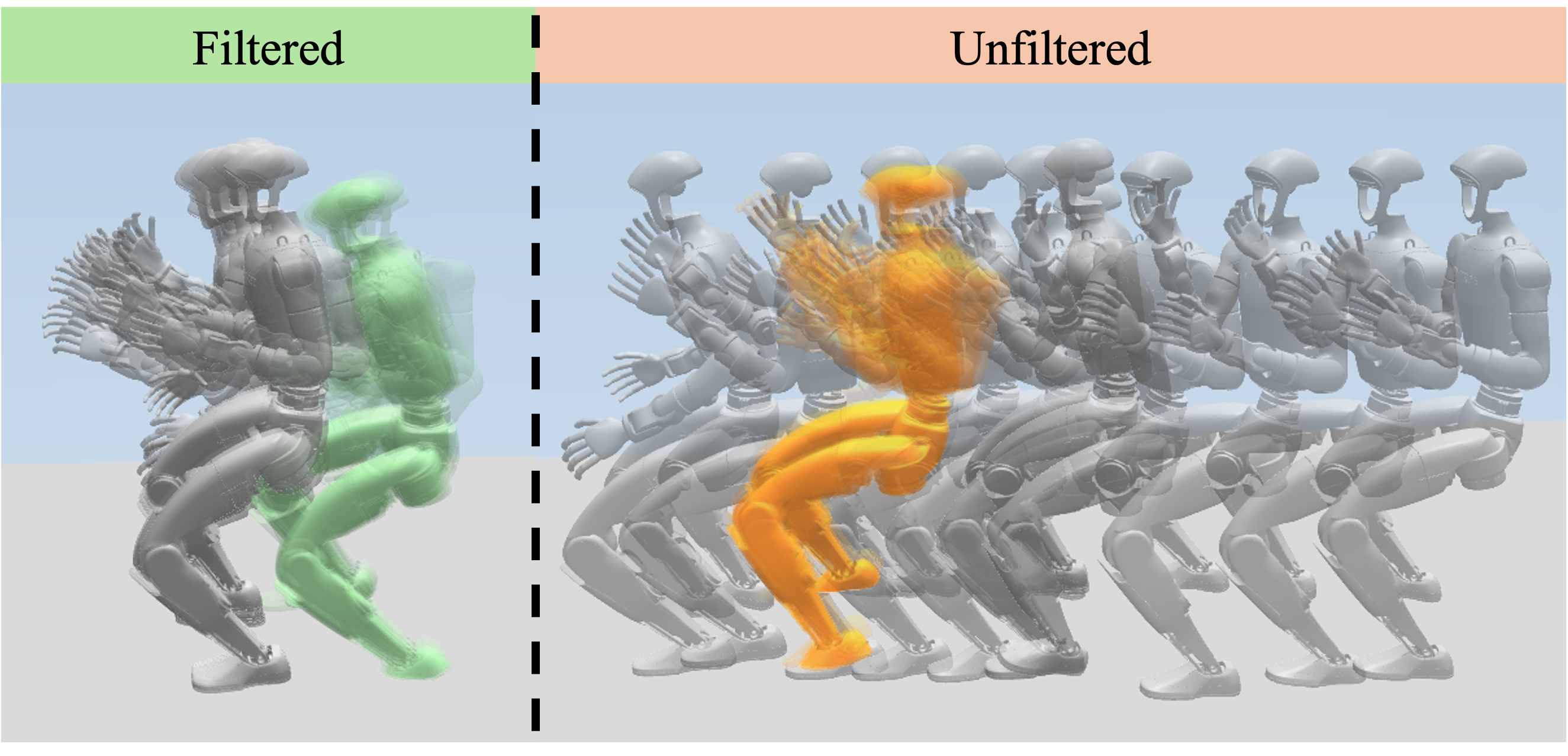}
\caption{\textbf{CoP filtering.} Filtered squat reference remains inside the support polygon; the original reference drifts.}
\label{fig:cop-filter}
\end{figure}

The $29$-DoF motion-tracking teacher follows recent controllers~\cite{beyondmimic2025, ze2025twist2, humanplus2024, exbody2_2024} and uses asymmetric actor-critic PPO~\cite{pinto2018asymmetric}.
Training uses a retargeted human motion dataset\footnote{Motion Data by Bones Studio - https://bones.studio/}~\cite{bones_seed_2026}.
Its actor receives an $11$-frame proprioceptive history and the full $29$-D reference joint angles from the current clip frame.
The critic also receives measured base linear velocity, reference root pose, key-body positions, and domain-randomization parameters, none of which are available to the deployed actor.
This teacher learns posture, reaching, squatting, and bilateral coordination.
However, its locomotion is tied to the motion data and does not reliably track arbitrary velocity commands, as shown in Table~\ref{tab:ablation-velocity}.
The retargeted data also contains squat frames that are dynamically infeasible for the real robot.
Before training, we correct them with a closed-form CBF projection on the static center-of-pressure (CoP) margin.
For quasi-static squat frames (root height below $0.65$~m and planar speed below $0.2$~m/s), let $h(q)$ be the signed CoM distance to the foot-contact box shrunk by $\delta=0.10$~m.
When $h<h_{\mathrm{tgt}}=0.04$~m, we project in a 7-D correction subspace of bilateral hip pitch, ankle pitch/roll, and waist pitch:
\begingroup
\setlength{\abovedisplayskip}{2pt}
\setlength{\belowdisplayskip}{4pt}
\begin{equation}
u^*=\frac{\max(0,h_{\mathrm{tgt}}-h(q))}{\|a(q)\|^2}a(q),\quad
a=E^\top[J^{\mathrm{cc}}_{c,xy}]^\top\nabla h,
\label{eq:cop-projection}
\end{equation}
\endgroup
where $J^{\mathrm{cc}}_{c,xy}$ is the contact-consistent CoM Jacobian under rigid double support.
We iterate this projection with backtracking, re-anchor the base at the mid-feet pose, and blend each corrected segment over 10 frames.
Figure~\ref{fig:cop-filter} shows one corrected squat clip.

\vspace{-5pt}
\subsection{Locomotion Teacher}
\label{sec:loco-teacher}
\vspace{-2pt}
The locomotion teacher controls the $15$-DoF legs-and-waist slice on flat terrain.
Curriculum-blended motion samples drive the arms during training, exposing the policy to the corresponding CoM shifts before distillation.
The actor observes commanded velocity, projected gravity, base angular velocity, joint state, last action, and a 4-D gait-phase block.
The critic also receives the measured base linear velocity.
The rewards combine velocity tracking with gait and stance terms.
Linear and angular velocity tracking use exponential errors with ranges of $1$~m/s and approximately $\pi$~rad/s.
A phase-contact term uses a $0.6$ stance ratio, while swing-foot clearance, flat-foot contact, foot-slip, and action-rate terms regularize the resulting gait.
\vspace{-5pt}
\subsection{Fall-recovery Teacher}
\label{sec:amp-teacher}

A full-body $29$-DoF Adversarial Motion Prior (AMP) teacher learns fall recovery~\cite{peng2021amp,amp_mjlab}.
Episodes start from random frames of retargeted motion, including ground poses from one fall-and-get-up clip.
The discriminator data contains this clip together with 17 walking, jogging, and turning clips.
The teacher does not copy the get-up motion frame by frame.
Under the style prior, RL learns recovery from fallen initializations, and a small torso-anchor reward encourages command tracking after the robot stands.
The PPO reward is $0.25r_{\mathrm{disc}}+0.75r_{\mathrm{task}}$.
Up to 40\% of environments use delayed fallen resets to retain ground-to-upright transitions in the training distribution.

\begin{figure*}[t]
\centering
\includegraphics[width=\textwidth]{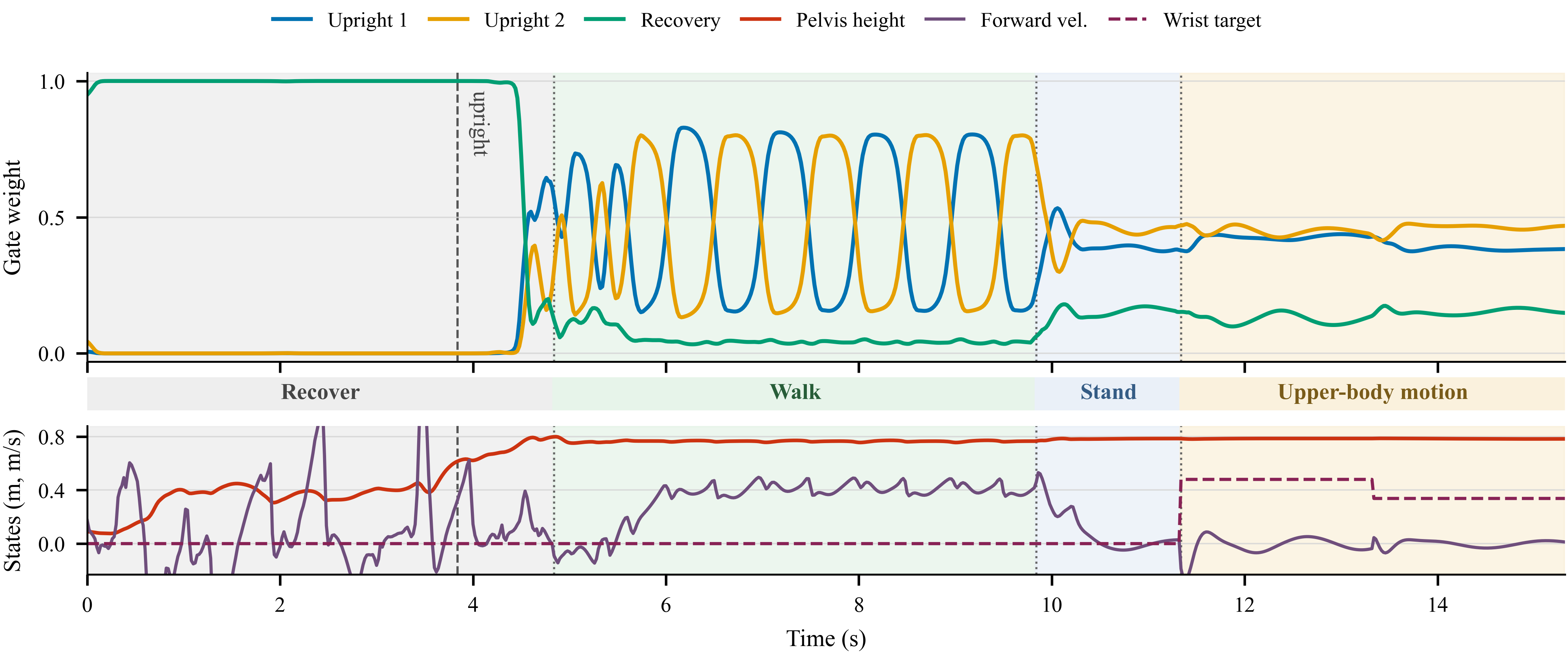}
\caption{\textbf{Expert routing.} Recovery dominates get-up; upright experts alternate in walking and share standing motions.}
\label{fig:routing-timeline}
\end{figure*}

\vspace{-5pt}
\subsection{Context-Conditioned Behavior Composition}
\label{sec:gated-kl}
The teachers provide complementary supervision at different levels: WBC and AMP cover the full action space in their respective regimes, while locomotion specializes in the body slice under velocity commands.
We combine their targets according to both action coverage and control regime.
Action slices determine which joints a teacher supervises, and context weights determine when its target is active.
We query them with a shared simulator state while preserving their native inputs: a dense motion reference for WBC, twist and gait phase for locomotion, and four-frame proprioception for AMP.
Pelvis-frame forward kinematics converts each WBC reference into the sparse commands $z,p_L^P,p_R^P$; velocity commands are sampled independently in $[-1,1]$ and resampled every 3--8~s, with 10\% of environments standing.
The student receives only the resulting 10-D command and its proprioceptive history.

Let $B=a_{0:15}$ and $A=a_{15:29}$ denote the body and arm slices, and let $D_j^S=D_{\mathrm{KL}}(\pi_\theta^S\|\pi_j^S)$ be the KL averaged over slice $S$.
We use context $\mathbf{x}_t=(\|c_t^{\mathrm{vel}}\|,r_t)$, where $r_t=\mathrm{recover}_t$, $m_t=1-r_t$, and $\alpha=\sigma((\|c_t^{\mathrm{vel}}\|-0.1)/0.02)$.
The four HANDOFF routes form one distillation loss:
\begin{align}
\mathcal{L}_{\mathrm{distill}}
&=m_t\lambda_B\!\left[(1-\alpha)D_{\mathrm{wbc}}^B+\alpha D_{\mathrm{loco}}^B\right]
 +m_t\lambda_A D_{\mathrm{wbc}}^A \nonumber\\
&\quad+r_t\lambda_{\mathrm{AMP}}D_{\mathrm{amp}}^{A\cup B}.
\label{eq:handoff-routes}
\end{align}
For upright states, $\alpha$ shifts body supervision from WBC to locomotion as commanded velocity increases, while WBC continues to supervise the arms.
In recovery states, AMP replaces both routes and supervises the full action.
For diagonal-Gaussian teachers, overlapping routes produce a precision-weighted geometric mean on each supervised action dimension.

To retain capacity for the three regimes within one controller, the student uses a soft MoE.
On non-recovery samples, a standard load-balancing loss $\mathcal{L}_{\mathrm{LB}}$~\cite{shazeer2017moe} balances the renormalized WBC and locomotion weights.
When recovery is active, $\mathcal{L}_{\mathrm{R}}$ pins one expert to recovery without labeling the two upright experts.
Figure~\ref{fig:routing-timeline} visualizes the learned routing weights across recovery, locomotion, and standing upper-body commands.
The complete student objective is
\begin{equation}
\mathcal{L}=\mathcal{L}_{\mathrm{PPO}}+\mathcal{L}_{\mathrm{distill}}
+\beta_{\mathrm{LB}}\mathcal{L}_{\mathrm{LB}}+\beta_{\mathrm{R}}\mathcal{L}_{\mathrm{R}}.
\end{equation}

\vspace{-5pt}
\subsection{Whole-Body Stability Shaping}
\label{sec:stability-shaping}
We add shared balance rewards~\cite{poddar2026embedding} to WBC, locomotion, and the student; AMP retains its recovery objective.
The rewards use the mass-weighted whole-body CoM and active-foot contacts.
CoM and capture-point terms reward staying inside the support region, where $\mathrm{CP}=\mathrm{CoM}_{xy}+\dot{\mathrm{CoM}}_{xy}/\sqrt{g/h_{\mathrm{CoM}}}$.
An ankle-hip-step term rewards ankle torque opposing CoM drift, hip motion once ankle authority saturates, and swing-foot motion toward the capture point once both saturate.
Actuator torques are normalized by their joint limits, and CoP saturation against a $\mu=0.7$ friction cone determines when ankle and hip authority are exhausted.
During training, we approximate the support polygon by the axis-aligned bounding box of active foot contacts, using force thresholds of 20~N for the CoM term and 50~N for the capture-point term.
This approximation is exact in double stance and avoids convex-hull construction during batched training.
For recovery states, active hand contacts extend the support region used by the CoM term.
If a wall or table contact supports the body, the capture-point term receives full reward rather than penalizing a point outside the feet-only polygon.
The same balance objective therefore covers foot-supported and braced whole-body recovery.
Penalties on changes in whole-body linear and angular momentum suppress abrupt corrections.
These terms shape upright balance, while AMP remains responsible for ground-to-upright recovery.

\subsection{Training setup}
All policies use asymmetric actor-critic PPO in mjlab with Rsl-rl.
One iteration collects 24 steps from 4,096 parallel environments ($\approx98$K transitions), followed by five Adam epochs over four mini-batches.
We use learning rate $10^{-3}$, PPO clip $0.2$, entropy coefficient $0.005$, $\gamma=0.99$, GAE $\lambda=0.95$, and gradient-norm clip $1.0$.
The teacher actors and critics are $[512,256,128]$ ELU MLPs.
The student uses a $[512,512,256,128]$ backbone with Swish, LayerNorm, and the 64-D temporal latent used by the MoE.
The MoE gate is a $[128,64]$ MLP over this latent, and each $[256,128]$ expert predicts a 29-D action mean.
At every step, the gate blends all outputs under a shared log-standard-deviation.
Only recovery is explicitly pinned; the other experts divide upright behavior without labels.
Soft routing avoids hard top-$k$ discontinuities and retains one deployed controller.
During distillation, $(\lambda_B,\lambda_A,\lambda_{\mathrm{AMP}})$ are cosine-annealed from $(0.4,0.1,0.4)$ to $(0.2,0.05,0.2)$, with $\beta_{\mathrm{LB}}=0.01$ and $\beta_{\mathrm{R}}=0.5$.
The smaller arm weight reduces interference with locomotion stability.
The WBC teacher and student train for up to 30K updates, the locomotion teacher for 20K, and AMP for 100K.
We evaluate the final checkpoints.
Domain randomization covers base mass, foot friction, PD gains, motor strength, encoder bias, hand forces, and actuation delay.
These parameters are hidden from the actor.

\subsection{Agentic planner}
\label{sec:planning}
\begin{figure*}[t]
\centering
\includegraphics[width=\textwidth]{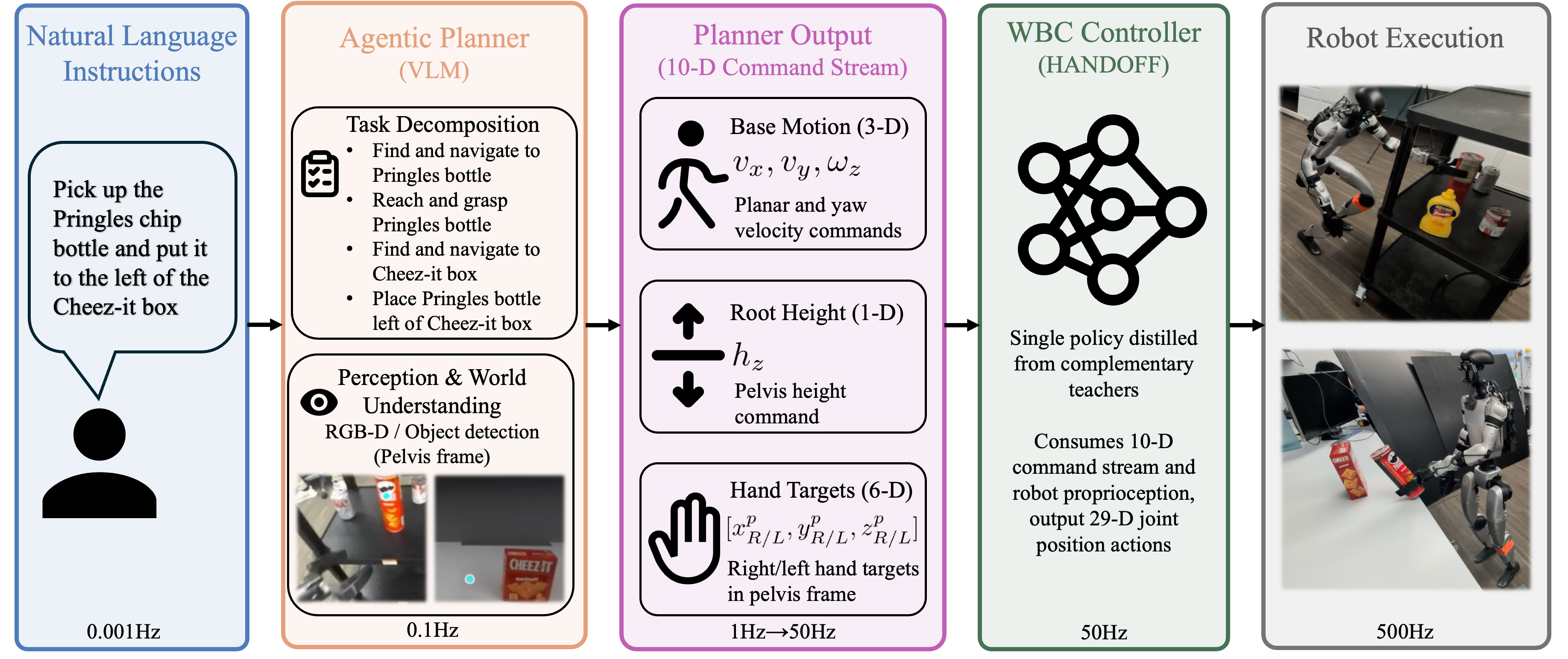}
\caption{\textbf{Agentic deployment pipeline.} Perception and planning produce the 10-D command consumed by HANDOFF.}
\label{fig:deployment}
\end{figure*}

Any planner that emits the 10-D command can drive the controller.
For the hardware experiments, we use the loco-manipulation stack in Fig.~\ref{fig:deployment}.
A high-level reasoner divides a natural-language instruction into atomic tasks through regex parsing, with an LLM fallback when no pattern matches.
A VLM projects predicted 2D points and bounding boxes into the RGB-D point cloud.
A waypoint tracker then converts the pelvis-frame waypoints into $(v_x,v_y,\omega_z)$.
Near the target, a skill selector produces root height $z$ and bilateral wrist targets $p_L^P,p_R^P$.
A kinematic correction aligns each gripper horizontally with the grasping surface.
The reasoner runs at 0.001~Hz, the VLM at 0.1~Hz, and the skill selector at 1--50~Hz.
HANDOFF runs at 50~Hz, and its 29-DoF actions are tracked on hardware at 500~Hz.

\section{Experiments}
\label{sec:experiments}

We run all experiments on a 29-DoF Unitree G1 with an external payload (Jetson Thor + Dex1-1 grippers) in mjlab/MuJoCo~\cite{zakka2026mjlablightweightframeworkgpuaccelerated}, and train every policy with Rsl-rl~\cite{schwarke2025rsl}.
We measure velocity tracking, manipulation workspace, fall recovery, and task resumption.
The experiments also isolate the components of the training architecture, compare controllers adapted to the same command interface, and test complete loco-manipulation tasks through the 10-D command.

For velocity tracking, we sweep per-axis commands along $[-1,1]$ and report $|\Delta v_\cdot| = \mathbb{E}_t[|v_\cdot^{\mathrm{cmd}}(t)-v_\cdot^{\mathrm{base}}(t)|]$.
For manipulation, we sample bilateral wrist targets uniformly inside $[-0.6,0.6]^3$~m in the pelvis frame, with a 1~s warm-up, 2~s settle, and 4~s measurement window.
We define \emph{feasibility} as the fraction of trials with no fall and less than 25~cm pelvis drift.
\emph{Robust WS} is the convex-hull volume of targets tracked within 15~cm, multiplied by feasibility, in the task-relevant forward half-space $x\geq0$.
Each velocity sweep uses 20 commands per axis.
Table~\ref{tab:ablation-velocity} uses 2,000 discovery and 400 accuracy targets per controller.
For the matched comparison, we select each adapted baseline on a separate validation set and evaluate it with 600 discovery and 250 accuracy targets per mirrored or independent mode.
All controller variants use the same command sweeps and paired target sets.

\begin{table}[!b]
\centering
\footnotesize
\caption{\textbf{Controller evaluation.} Panel (a) isolates the stages of behavior composition; panel (b) compares controllers under the matched adapted contract.}
\label{tab:quant-combined}
\begin{minipage}[t]{\linewidth}
\centering
\subcaption{Progressive behavior composition.}
\label{tab:ablation-velocity}
\resizebox{\linewidth}{!}{%
\renewcommand{\arraystretch}{1.25}%
\begin{tabular}[t]{l cccc}
\toprule
\textbf{Ablation} & $|\Delta v_x|$ & $|\Delta v_y|$ & $|\Delta \omega_z|$ & \textbf{Robust WS} \\
                  & \multicolumn{3}{c}{(m/s, rad/s) $\downarrow$} & (m$^3$) $\uparrow$ \\
\midrule
Direct                            & 0.29          & 0.43          & 0.08          & 0.20 \\
+ Dual teacher                    & 0.14          & 0.25          & 0.09          & \textbf{0.31} \\
+ RandCmd                         & 0.14          & \textbf{0.13} & 0.05          & 0.29 \\
\textbf{+ Split KL + MoE (Ours)}      & \textbf{0.07} & 0.14          & \textbf{0.04} & 0.27 \\
\textbf{+ AMP recovery (Ours + Rec.)} & \textbf{0.07} & 0.15          & 0.05          & 0.26 \\
\textbf{+ Stability (Ours + Stab.)}   & \textbf{0.07} & 0.18          & \textbf{0.04} & \textbf{0.31} \\
\textbf{+ Stab. + Rec. (Ours + Stab. + Rec.)} & \textbf{0.06} & 0.18 & 0.06 & 0.277 \\
\bottomrule
\end{tabular}}
\end{minipage}
\par\medskip
\begin{minipage}[t]{\linewidth}
\centering
\subcaption{Matched adapted-contract comparison.}
\label{tab:main-quant}
\resizebox{\linewidth}{!}{%
\renewcommand{\arraystretch}{1.05}%
\begin{tabular}[t]{l ccc c c}
\toprule
\textbf{Method} & $|\Delta v_x|$ & $|\Delta v_y|$ & $|\Delta \omega_z|$ & \textbf{Feas.} & \textbf{Robust WS} \\
                & \multicolumn{3}{c}{(m/s, rad/s) $\downarrow$} & (\%) $\uparrow$ & (m$^3$) $\uparrow$ \\
\midrule
FALCON~\cite{falcon2026}    & 0.07          & \textbf{0.10} & 0.07          & 26.1 & 0.033 \\
OpenHomie~\cite{homie2025}  & 0.06          & 0.12          & 0.10          & 72.3 & 0.098 \\
AMO~\cite{amo2025}          & 0.06          & 0.19          & \textbf{0.01} & 84.7 & 0.215 \\
SONIC~\cite{sonic2025}      & \textbf{0.03} & 0.15          & 0.02          & \textbf{96.2} & 0.265 \\
\textbf{Ours + Stab. + Rec.}& 0.06          & 0.18          & 0.06          & 93.0 & \textbf{0.277} \\
\bottomrule
\end{tabular}}
\end{minipage}
\vspace{-10pt}
\end{table}

\subsection{Ablation studies}
\label{sec:ablation}
Table~\ref{tab:ablation-velocity} starts with direct distillation of the motion-tracking teacher and adds one specialist or composition mechanism at a time.
Each row is one policy with the same 10-D interface.
The locomotion teacher gives the largest initial velocity gain.
Randomized commands then close the lateral-velocity gap, and action-sliced KL with a soft MoE closes the remaining $v_x$ gap.
Two-teacher ablations use two experts.
The recovery variants add a third pinned expert and recovery KL term, while the stability variants change rewards but not the teacher-student architecture.
Gate probes on deployed observation streams show the learned specialization.
The recovery expert receives 92\% of the gate mass during get-up and 5\% during walking, while upright motion mixes the two non-recovery experts with weights of 0.40--0.49 each.
Adding recovery expands the hull toward more extreme targets but reduces pointwise feasibility.

\begin{figure*}[t]
\centering
\includegraphics[width=\textwidth]{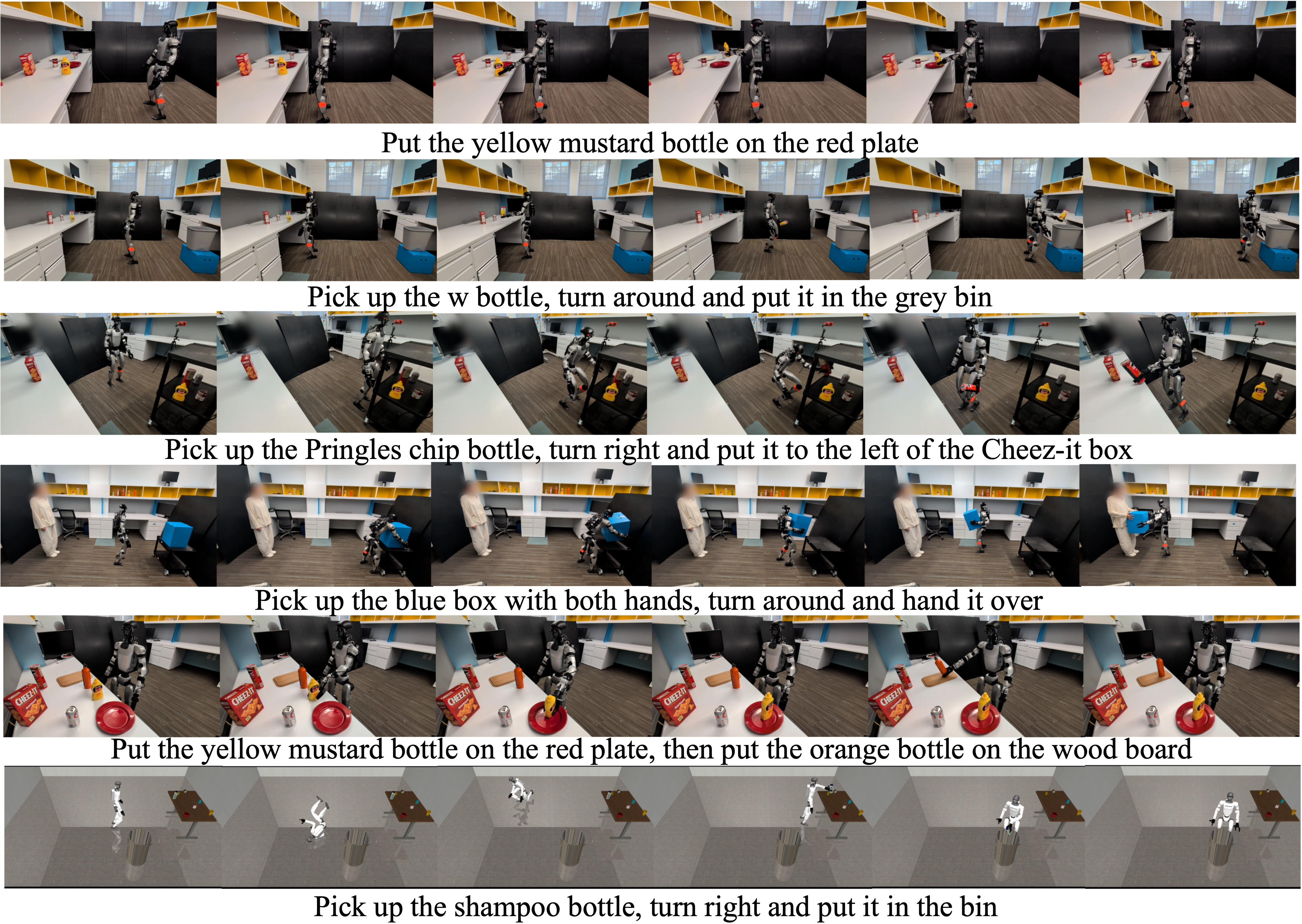}
\caption{\textbf{Hardware and simulated deployments.} The same 10-D controller executes six VLM-commanded task sequences without task-specific fine-tuning.}
\label{fig:experiment-snapshots}
\end{figure*}

\subsection{Comparison with whole-body controllers}
\label{sec:sota}
\begin{figure}[!t]
\centering
\includegraphics[width=\columnwidth]{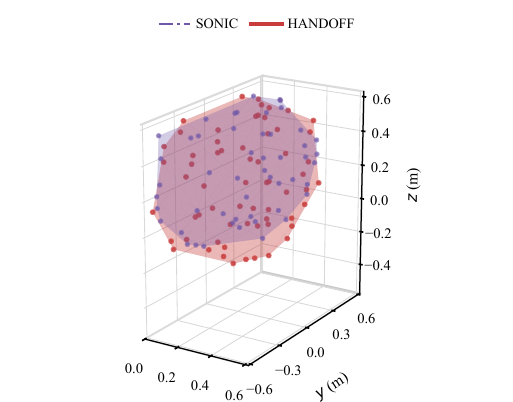}
\caption{\textbf{Bilateral wrist workspace.} Feasible target hulls are shown in the pelvis frame under the validation-selected protocol.}
\label{fig:workspace-hulls}
\vspace{-25pt}
\end{figure}

Table~\ref{tab:main-quant} uses a common \emph{adapted contract} because the prior controllers accept different arm interfaces.
The same differential-IK bridge~\cite{mink} maps bilateral wrist targets to each baseline's native arm commands, while its controller remains responsible for walking and balance.
Figure~\ref{fig:workspace-hulls} visualizes the workspace hulls combined across mirrored and independent modes.
HANDOFF covers $0.297$~m$^3$, giving $0.277$~m$^3$ Robust WS; SONIC has higher feasibility at 96.2\% but a smaller $0.275$~m$^3$ hull and $0.265$~m$^3$ Robust WS.
Conditioning on targets where the IK bridge converges below a 2~cm residual, AMO, OpenHomie, and SONIC execute 87--92\%, while FALCON executes 63\%.

\begin{table}[htpb]
\centering
\footnotesize
\vspace{-10pt}
\caption{\textbf{Task-level success and recovery.} Values are percentages except mean recovery time.}
\label{tab:rebuttal-evals}
\begin{minipage}[t]{\linewidth}
\centering
\subcaption{Cumulative task-stage success.}
\label{tab:task-success}
\setlength{\tabcolsep}{6pt}
\resizebox{\linewidth}{!}{%
\begin{tabular}{lccccc}
\toprule
Task & Approach & Grasp & Carry & Place/HO & Success \\
\midrule
Pick--place & 100 & 100 & --- & 99.8 & 99.8 \\
Bilateral pick--place & 100 & 100 & --- & 100 & \textbf{100} \\
Bimanual handover & 100 & 77.6 & 77.6 & 77.6 & 77.6 \\
Squat pick & 99.8 & 95.6 & 71.0 & 67.2 & 67.2 \\
Pick--transport--place & 99.8 & 99.8 & 72.4 & 72.2 & 72.2 \\
\bottomrule
\end{tabular}}
\end{minipage}
\par\smallskip
\begin{minipage}[t]{\linewidth}
\centering
\subcaption{Fallen-state recovery and strong-push robustness.}
\label{tab:recovery}
\setlength{\tabcolsep}{6pt}
\resizebox{\linewidth}{!}{%
\begin{tabular}{lcccc}
\toprule
Variant & Recovery & $\bar t_{\rm rec}$ (s) & Resume & Upright@6$\times$ \\
\midrule
Ours & 0 & --- & --- & 41 \\
Ours + Stab. & 0 & --- & --- & 55 \\
Ours + Rec. & 49.8 & 1.5 & 100 & 74 \\
Ours + Stab. + Rec. & \textbf{89.6} & 3.2 & \textbf{100} & \textbf{95} \\
StrongPush & 0 & --- & --- & 66 \\
\bottomrule
\end{tabular}}
\end{minipage}
\vspace{-10pt}
\end{table}

\subsection{Recovery and task resumption}
\label{sec:recovery-eval}
All variants receive random base-velocity impulses every 1--3~s, up to 0.5~m/s per horizontal axis.
They remain upright at four times this magnitude; Upright@6$\times$ extends the disturbance-rejection test.
Table~\ref{tab:recovery} instead starts each policy from 500 paired, physics-generated fallen states.
The stability rewards improve both recovery and the six-times-push result.
Every successful recovery resumes a commanded 0.4~m/s walk with at most 0.05~m/s error.
The benchmark reports standing and command resumption as separate metrics, with recovery time computed over successful trials.
Paired initial states and disturbance sequences keep the comparison across variants fixed.
StrongPush is trained directly with fall-inducing impulses at six times the training magnitude.
It records no recoveries in 500 trials, along with 47\% shorter episodes and 14\% higher velocity error.
Falls terminate these episodes, so its training data do not contain ground-to-upright transitions.
The AMP teacher supplies these transitions as a separate recovery regime.

\subsection{Task-level contract and hardware deployment}
\label{sec:deployment}
Our VLM+depth stack estimates goals in the robot frame and re-queries at close range without a global scene map.
Deterministic tools provide waypoint tracking, path planning, skill selection, model-based wrist-orientation correction, and discrete gripper control on top of the controller's 3-D wrist-position commands.
The VLM grounds targets to 1.2--1.6~cm error after re-query.
We inject that measured error into 500 paired, randomized object/goal placements for each of five tasks while retaining the deployed downstream tools and 10-D controller contract; grasps are kinematic within the gripper's 12~cm jaw width.
In Table~\ref{tab:task-success}, Place/HO denotes the terminal placement or handover stage.
Across 2,500 trials, the approach stage succeeds in at least 99.8\%, no trial falls, and total task success ranges from 67.2\% to 100\% (Table~\ref{tab:task-success}).
Because the entries are cumulative, each column includes all preceding stages.
Bilateral pick-and-place remains at 100\% through placement, while bimanual trials that complete grasp retain the same success rate through carry and handover.

We deploy the same controller on a real Unitree G1 with a ZED-M RGB-D camera and Jetson Thor, using both local VLM and ChatGPT APIs.
Figure~\ref{fig:experiment-snapshots} shows pick-and-place, pick-transport-place, squat-pick, bimanual handover, bilateral pick-and-place, and task continuation after recovery; Table~\ref{tab:task-success} reports aggregate task statistics from simulation.
In these hardware runs, one unchanged low-level policy executes all six command sequences through the full planning stack.
The high-level stack changes only the 10-D command and does not select among low-level controllers.
Hardware operation uses command clipping, wrist-target workspace bounds, CoP filtering, and an operator e-stop.

\section{Conclusion}

In this paper, we introduced HANDOFF, which distills specialists with different objectives and action coverage into one humanoid whole-body controller.
Its context-conditioned, action-sliced objective blends motion-tracking and locomotion supervision over the body, keeps the arms under whole-body motion supervision, and assigns the full action to the recovery teacher only during recovery.
A soft mixture of experts combines these regimes without switching controllers at deployment.
The policy matches state-of-the-art velocity tracking, obtains the largest robust workspace in a matched adapted-interface comparison, and recovers from the ground before resuming its commanded walk.
The recovery ablation also separates disturbance rejection from recovery: only the teacher that observes ground-to-upright transitions learns to stand again.
All behaviors use one 10-D command, allowing the controller to operate below different planning systems.
We evaluate this interface with randomized task trials and VLM-driven hardware experiments.
Future work will extend this composition scheme to additional specialists and test whether they can be added without changing the 10-D command or degrading behaviors already in the controller.

\balance
{\small
\bibliographystyle{IEEEtran}
\bibliography{references}
}

\end{document}